\documentclass[11pt]{article}

\usepackage[final]{acl}

\usepackage{times}
\usepackage{latexsym}
\usepackage{ascmac}
\usepackage{amsfonts}
\usepackage{amsmath}
\usepackage{bxcoloremoji}
\usepackage{soul}
\usepackage{url}
\usepackage{fancyhdr}

\usepackage{listings}
\usepackage[T1]{fontenc}

\usepackage[utf8]{inputenc}

\usepackage{microtype}

\usepackage{inconsolata}

\usepackage{graphicx}

\usepackage{multirow}
\usepackage{tabularx}
\usepackage{boites,boites_exemples}

\usepackage{algorithm}
\usepackage{algorithmicx}
\usepackage[noend]{algpseudocode}
\algnewcommand\algorithmicinput{{\bfseries\gtfamily Input}}%
\algnewcommand\algorithmicoutput{{\bfseries\gtfamily Output}}%
\algnewcommand\AlgInput{\item[\algorithmicinput]}%
\algnewcommand\AlgOutput{\item[\algorithmicoutput]}%
\algrenewcommand\Return{\State\textbf{return} }%

\title{Natural Language Translation of Formal Proofs through Informalization of Proof Steps and Recursive Summarization along Proof Structure}

\author{%
  Seiji Hattrori${}^{1}$ Takuya Matsuzaki${}^{1}$ Makoto Fujiwara${}^{1}$\\
${}^{1}$Tokyo University of Science\\
\ \ \texttt{1424521@ed.tus.ac.jp}
\ \ \texttt{matuzaki@rs.tus.ac.jp}
\ \ \texttt{makotofujiwara@rs.tus.ac.jp}
}

\begin{document}
\maketitle
\begin{abstract}
This paper proposes a natural language translation method for machine-verifiable formal proofs that leverages the informalization 
(verbalization of formal language proof steps) and summarization capabilities of LLMs. For evaluation, it was applied to formal proof data created in accordance with natural language proofs taken from an undergraduate-level textbook, and the quality of the generated natural language proofs was analyzed in comparison with the original natural language proofs. Furthermore, we will demonstrate that this method can output highly readable and accurate natural language proofs by applying it to existing formal proof library of the Lean proof assistant.
\end{abstract}
\begin{figure*}[t]
 \begin{center}
  \includegraphics[width=13cm]{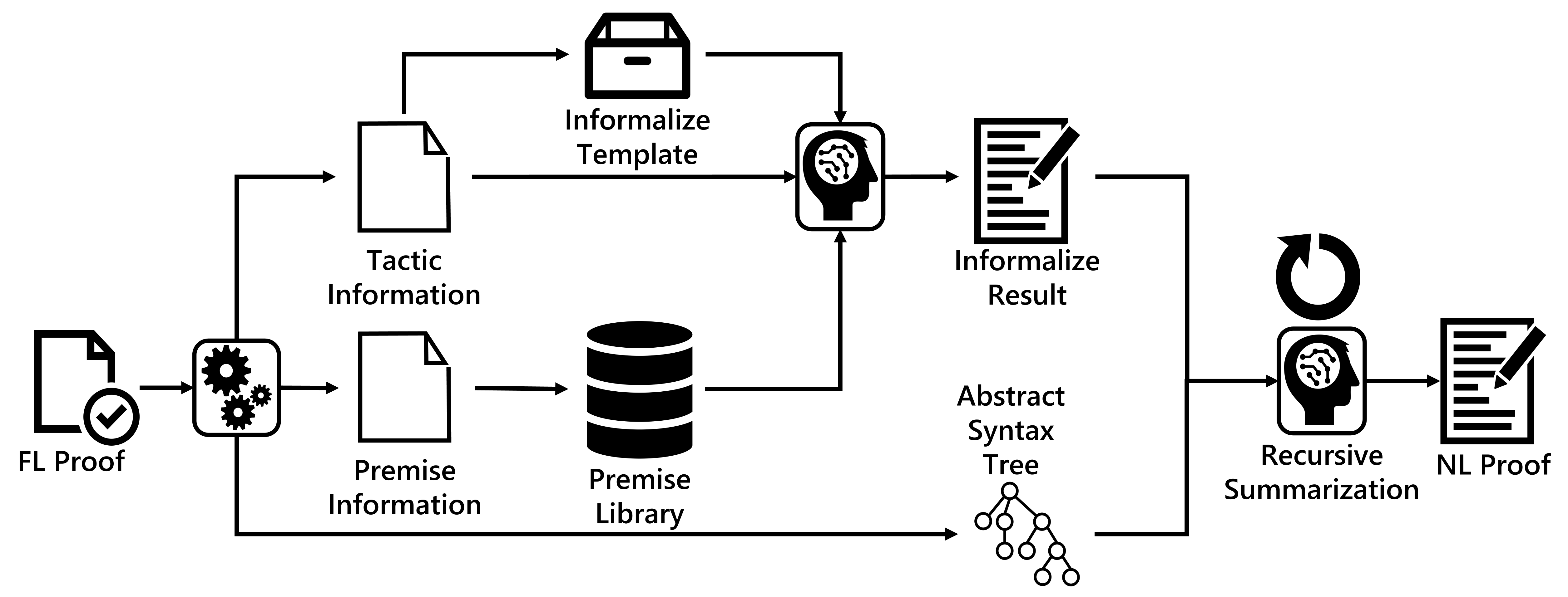}
  \caption{Method Overview}
  \label{fig:method}
 \end{center}
\end{figure*}
\section{Introduction}
Mathematical proofs written in a formal language that computers can verify are known as formal proofs. These are primarily written in languages used by theorem proving assistants such as Isabelle \cite{isabelle}, Rocq \cite{coq}, and Lean \cite{lean}.

Autoformalization is a technology that automatically converts human-written proofs (natural language proofs) into formal proofs. 
It reduces the burden of manually formalizing mathematical theorems. Moreover, in the domain of natural language theorem proving using large language models (LLMs) (e.g., \citealp{jiang2023draft}, \citealp{zhang2025leanabellproverposttrainingscalingformal}, \citealp{xin2025deepseekproverv}), autoformalization enables us to programmatically verify the logical validity of generated content, thereby contributing to the improvement of automated theorem provers.

In recent years, with the advancement of LLMs, research into autformalization as a machine translation task has become active (e.g.,  \citealp{li2024autoformalize}; \citealp{tarrach2024more}; \citealp{zhang2025formalizingcomplexmathematicalstatements}). However, challenges persist due to the lack of natural language proofs paired with formal proofs for model training. Consequently, its target remains primarily on high school level mathematics, such as miniF2F \cite{miniF2F}, where the emphasis is on arithmetic operations rather than mathematical reasoning.

This paper proposes an auto-\textit{informalization} method that translates formal proofs into natural language proofs, addressing the challenge of limited training data for autoformalization. To achieve this, we need to solve the following challenges:\\
\textbf{Generation of Natural Reference Expression} Mathematical proofs frequently refer to predefined concepts, theorems, and assumptions made in the course of a proof. In formal proofs, these references are made using arbitrarily named labels. However, simply using these labels during translation cannot produce natural reference expressions as those written by humans. \\
\textbf{Abstraction to Human-Readable Proofs} It is possible to mechanically translate each piece of a formal proof into corresponding natural language expression, as all operations in formal language have clear meanings. However, a natural language proof constructed by mechanically translating and concatenating individual pieces often results in an unnatural and incomprehensible output. To achieve a natural translation, it is essential to include information that is both necessary and sufficient for human readers by an appropriate abstraction.

To tackle the above problems, we combine the following three components:\\
\textbf{Construction of a Premise Library} Our method performs translation using a \textbf{premise library}, which is created by converting definitions and propositions within a formal mathematics library (Mathlib) into natural language descriptions in advance. These descriptions are included in the prompt for step-wise informalization to encourage the generation of natural reference expressions.\\
\textbf{Step-wise Informalization} Each step of a formal proof (consisting of commands called tactics) is translated into a natural language proof step. It is generated by a hybrid method that combines a rule-based approach using templates prepared for each tactic type, and slot-filling by an LLM that extracts corresponding information based on slot types that specify the content to be included.\\
\textbf{Summarization of Proof Steps} We generate a natural language proof by progressively summarizing the description of the proof steps, reflecting the structure of the formal proof.

The generated natural language proofs must accurately reflect the logical structure expressed in the formal proofs, and furthermore, must correctly capture the essential points of the proof to avoid logical leaps or breakdowns. To evaluate this, we prepared formal proof data by formalizing university undergraduate-level mathematical proofs, maintaining their proof structure as much as possible. We then applied the proposed method to this data and compared the generated output with the original natural language proofs to conduct a detailed evaluation. Additionally, we will present conversion examples for Mathlib proofs, which do not have corresponding natural language proofs, to demonstrate that high quality natural language proofs can be generated.

\section{Related Work}
This section introduces several recent studies on autoformalization and informalization that are particularly relevant to our research. 

Jiang et al. \shortcite{jiang2024multilingual} proposed a method to construct a large-scale dataset of pairs of formal proposition and natural language proposition by informalizing formal propositions written in Isabelle and Lean using GPT-4. While our proposed method also involves prompt-based informalization, it differs from Jiang et al.'s approach in that their method targets only formal \textit{propositions}, whereas our research focuses on formal \textit{proofs}.

Although prior work such as \citet{CHESTER1976261} and \citet{DBLP:conf/aaai/Holland-MinkleyBC99} has addressed the translation of formal proofs into natural language proofs, there has been relatively little research on approaches that utilize LLMs.
Gao et al. \shortcite{gao2024heraldnaturallanguageannotated} proposed a method for constructing high-quality paired datasets of formal and natural language proofs. Their approach uses static analysis tools to extract information such as code comments, similar propositions, and dependent propositions from Lean’s formal proof library, and then provides this information to the LLM as auxiliary input to facilitate the translation of formal proofs. In addition, they aim for high-quality step-wise informalization and natural language proof generation by manually creating explanatory sentences for each tactic’s logical structure and incorporating them into the prompt, thereby helping the LLM capture the relationship between proof steps and goals.

While our method is inspired by theirs, it differs in two ways. First, instead of guiding the LLM with explanatory text, we directly control the form of step-wise informalization by applying rule-based template generation. Second, whereas their method generates natural language proofs by concatenating step-wise results, our approach recursively summarizes these results according to the structure of the formal proof, producing text that is more readable and closer in style to human-written proofs.

\section{Theorem Proving Assistant}
A theorem proving assistant refers to software that provides a formal language for writing mathematical proofs and an automated verification system for proofs written using that formal language.

Lean \cite{lean} is one such proof assistant, offering a functional programming language based on dependent type theory. Similarly to Rocq and Isabelle, Lean provides interactive proof assistance through commands called "tactics." Furthermore, formal proofs can be created efficiently by leveraging Mathlib, a formal proof library built on Lean by its user community.

We use Lean 4 and Mathlib4, released in 2021 for the implementation of the proposed method and the creation of few-shot examples.

We explain the gist of Lean's formal proof using the proposition "the sum of two even numbers is an even number" and its proof.
A proof of this proposition is given as follows:
\begin{screen}
\parindent0pt
\fontsize{10pt}{11pt}\selectfont
If $a$ and $b$ are even numbers, then there exist some $r_1, r_2$ such that\\
\centerline{$a = r_1 + r_1, b = r_2 + r_2.$}\\
This means that\\
\centerline{$a + b = (r_1 + r_2) + (r_1 + r_2).$}\\
Therefore, $a + b$ is also an even number.
\label{ex:NL-Proof}
\end{screen}
\begin{figure}[t]
\begin{lstlisting}[mathescape]
theorem EvenAddEvenEqEven (a : $\mathbb{N}$) (b : $\mathbb{N}$) :
	(Even a $\wedge$ Even b) $\rightarrow$ Even (a + b) := by
	intro $\langle \langle$r1, h1$\rangle$ , $\langle$r2, h2$\rangle\rangle$
	have : a + b = (r1 + r2) + (r1 + r2) := by
		rw [Nat.add_assoc, Nat.add_comm r2,
			$\leftarrow$ Nat.add_assoc, $\leftarrow$ Nat.add_assoc]
		rw [h1, h2]
		rw [$\leftarrow$ Nat.add_assoc]
	exact $\langle$(r1 + r2), this$\rangle$
\end{lstlisting}
\caption{Formal proof of "the sum of two even numbers is an even number"}
\label{ex:FL-Proof}
\end{figure}
Figure \ref{ex:FL-Proof} presents a formalization of this proof.
In a Lean formal proof, we first declare a proposition to prove using the \texttt{theorem} command. In Figure \ref{ex:FL-Proof}, it is declared that for any natural numbers $a, b$, if $a$ and $b$ are even, then $a + b$ is also even (line 1-2).
Next, we introduce variables and hypotheses using the \texttt{intro} tactic (line 3). Since even numbers are defined as $ \text{Even}\ x \Leftrightarrow \exists r.\ x = r + r$, the following hypotheses are introduced: \\
\indent\texttt{r1 r2 : $\mathbb{N}$},\\
\indent\texttt{h1 : a = r1 + r1}, and \\
\indent\texttt{h2 : b = r2 + r2}.\\
The proof goal then becomes to prove \texttt{Even (a + b)}.
We then declare a sub-proposition to be proven using the \texttt{have} tactic (line 4), namely that $a + b = (r1 + r2) + (r1 + r2)$. Then, the proof goal temporarily changes from \texttt{Even (a + b)} to \texttt{a + b = (r1 + r2) + (r1 + r2)}, while the hypotheses remain the same. The subsequent three \texttt{rw} tactics (line 5-8) rewrite on the proof goal using the propositions in the library and hypotheses. The first tactic rewrites \texttt{$(r1 + r2) + (r1 + r2)$} to \texttt{$r1 + r1 + r2 + r2$} using the commutative and associative laws of addition. The next tactic rewrites \texttt{$a + b$} on the left-hand side of the goal using the two hypotheses \texttt{h1} and \texttt{h2} introduced by the \texttt{intro} tactic, and finally, by again using the associative law of addition, the left and right sides of the goal are made to have the same form, completing the proof of the sub-proposition.

Finally, the proof of the initially declared proposition is completed using the \texttt{exact} tactic (line 9). Here, \texttt{$r = r1 + r2$} is provided as evidence that \texttt{$a + b$} is even, thus completing the proof. Note that \texttt{this} given in the \texttt{exact} tactic refers to the proposition declared with the \texttt{have} command, which in this case refers to $a + b = (r1 + r2) + (r1 + r2)$.

In Lean, propositions are thus proven by repeatedly applying tactics and changing the state of the proof, i.e., the goal proposition and the established hypotheses.

\section{Method}

Our proposed method translates formal proofs written in Lean 4 syntax into natural language proofs through five stages: off-line generation of premise library (\ref{method-premise-library}), information extraction (\ref{method-extraction}), step-wise informalization (\ref{method-informalization}), dependency structure analysis (\ref{method-parse}), and summarization (\ref{method-summarization}). Figure \ref{fig:method} provides an overview of the method.

\subsection{Construction of Premise Library}\label{method-premise-library}
The premise library contains explanatory sentences for all theorems and definitions defined within Lean and Mathlib. This library is generated using the following procedure:
\begin{enumerate}
\item Information Extraction\newline
All theorems and definitions are extracted from Lean and Mathlib source codes (modules).
\item Dependency Analysis\newline
Based on the list of modules imported by each module, the dependencies between modules are expressed in terms of levels. The level value of each module indicates that it may depend on modules with smaller level values and does not depend on modules with larger level values.
\item Explanation Generation\newline
Starting from modules with smallest level values, an LLM is prompted to generate explanatory sentences for the theorems and definitions defined in that module. If a theorem or a definition depends on another definition, and its explanation is already available, the explanation is provided in the prompt. As few-shot examples, three other definitions and theorems belonging to the same module or the same field (e.g., linear algebra) are randomly selected and provided.
\end{enumerate}

For example, for the theorem representing the triangle inequality of absolute values ($|a + b| \leq |a| + |b|$), the output would be: ``This theorem states that in a linearly ordered additive group, the absolute value of the sum of the two elements is less than or equal to the sum of their absolute values, embodying the triangle inequality.''

This library is used in the subsequent informalization (\ref{method-informalization}) and summarization (\ref{method-summarization}) steps.

\subsection{Formal Proof Information Extraction}\label{method-extraction}
Various pieces of information from the Lean formal proof data are extracted for use in informalization, dependency structure analysis, and summarization steps. We extend LeanDojo \cite{leandojo}'s data extraction tool and use it for this purpose.
We extract three types of data:
\begin{itemize}
\item \textbf{Tactic Information} holds the type and the arguments of each tactic applied in a formal proof and the changes in the proof state before and after the tactic's application.
\item \textbf{Premise Information} contains information such as the names and types of theorems and definitions referenced in the formal proof.
\item \textbf{Abstract Syntax Tree (AST)} represents the dependencies between tactics, variables, and other elements contained in a formal proof. We obtain it from the internal data structure of the Lean system.
\end{itemize}
\subsection{Informalization of each proof step}\label{method-informalization}
\begin{figure}[t]
 \centering
 \includegraphics[width=7.5cm]{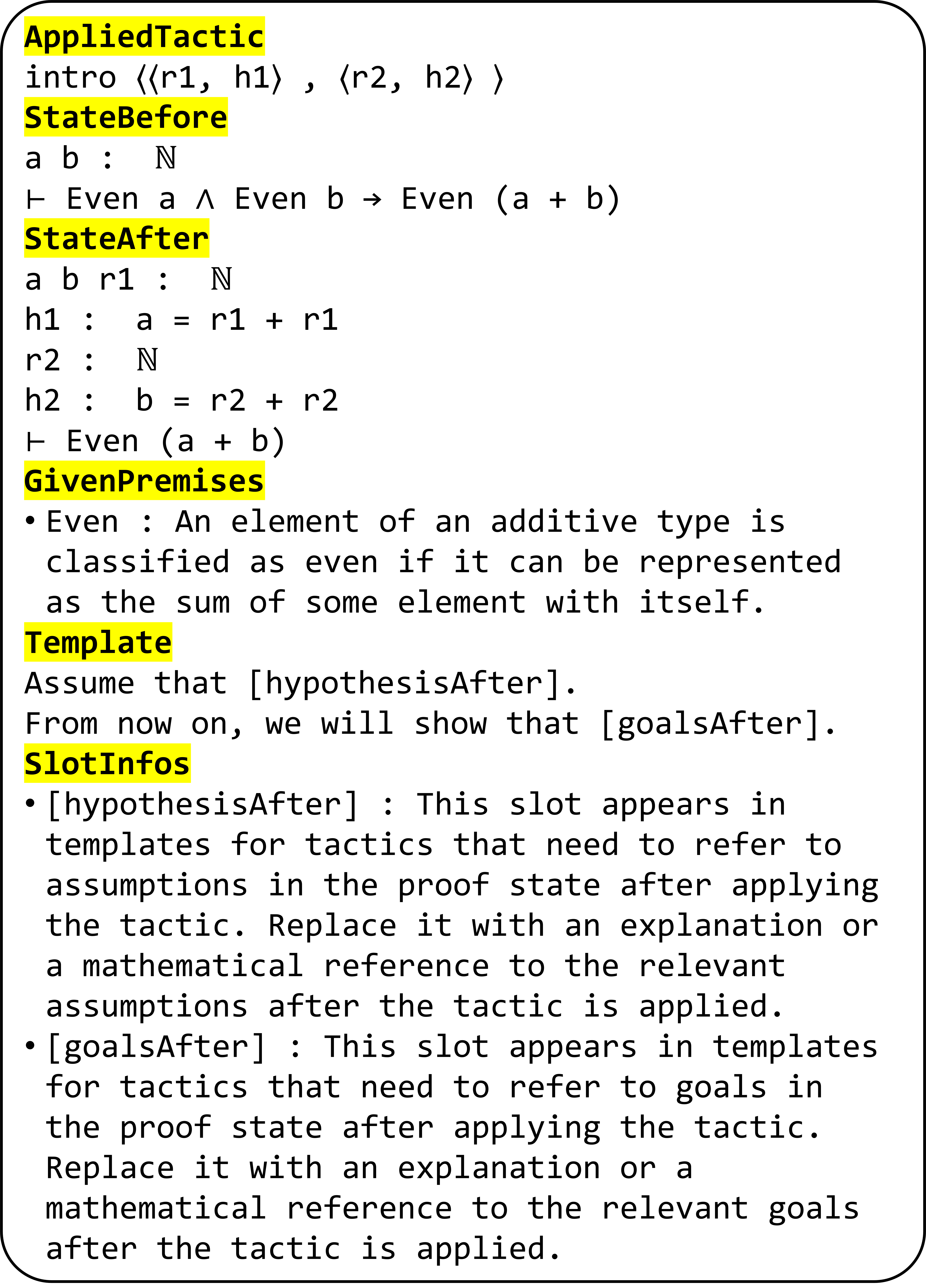}
 \caption{Informalize task input example\label{fig:input-example}}
\end{figure}
Using the tactic information and premise information extracted in the previous step, we generate explanatory sentences for the operations of each proof step in a formal proof by using an LLM. The main part of the prompt used is shown in Figure \ref{fig:input-example} and the full prompt is shown in Appendix B.

We first extract a tactic applied in a proof step from the tactic information, as well as the proof state before and after its application (all expressed in the form of hypotheses $\vdash$ goal).

Next, an appropriate generation template is retrieved. One or more template are manually prepared for each tactic; each template corresponds to different usage of the tactic.

Let's take the \texttt{rw} tactic used in lines 5 and 7 of Figure \ref{ex:FL-Proof} as an example. In the case of the \texttt{rw} tactic, templates are differentiated by whether it rewrites the goal, an existing hypothesis, or both, and whether the rewriting uses only hypotheses, only theorems, or both hypotheses and theorems. The \texttt{rw} tactic on line 5 rewrites the proof goal using two theorems, \texttt{Nat.add\_assoc} and \texttt{Nat.add\_comm}. The \texttt{rw} tactic on line 7 rewrites the proof goal using two previously introduced hypotheses, \texttt{h1} and \texttt{h2}. The template retrieval algorithm uses this information to proceed through the template branches and select the most appropriate template. In this example, the following templates are selected respectively for the two \texttt{rw} tactics:
\begin{screen}
	\parindent0pt
	\fontsize{8pt}{11pt}\selectfont
 	By using [theorems], [goalsBefore] becomes [goalsAfter].\\
 	By using [assumptions], [goalsBefore] becomes [goalsAfter].
\end{screen}
The templates include several slots, such as [theorems], [assumptions], [goalsBefore], and [goalsAfter]. The LLM is instructed to fill these slots so that they explain "theorems used during tactic application," "assumptions used during tactic application," "the proof goal before tactic application," and "the proof goal after tactic application," respectively. 
These explanations of the slots are provided in the informalization prompt along with the template itself.

Next, explanatory sentences corresponding to the theorems and definitions used in the applied tactic are retrieved from the premise library. For instance, in line 5 of Figure \ref{ex:FL-Proof}, which uses \texttt{rw [Nat.add\_assoc, Nat.add\_comm r2, ...]}, two theorems are used: \texttt{Nat.add\_assoc} and \texttt{Nat.add\_comm}.
Therefore, the premise library is searched based on the modules where these theorems are defined, and their respective explanatory sentences are retrieved.

Next, few-shot examples of the step-wise informalization are obtained from a manually created list.
Few-shot examples are created for each type of tactic operation defined in Lean and Mathlib. Each example consists of the applied tactic, the proof state before and after its application, and the resulting operation explanation sentence. The tactic and proof state of the few-shot examples are extracted from self-created formal proofs and Mathlib formal proofs. The corresponding output examples were all manually created with the intention of being natural as sentences included in a proof and clearly indicating to the LLM what operation the given tactic corresponds to and which part of the proof state to focus on for output. The formats of the example sentences adhere to one of the templates.

Finally, the information obtained through the preceding steps is provided to the LLM along with a prompt, and natural language explanations corresponding to each proof step are generated.

\subsection{Dependency Analysis of Proof Steps}\label{method-parse}

In this step, we analyze the dependencies between tactics using the abstract syntax tree extracted in \ref{method-extraction} and associate the operation explanation sentences corresponding to each proof step to the nodes of the tree structure that expresses these dependencies. This tree structure has the proposition being proven as its root and the explanation sentences of each proof step as its descendants. Furthermore, when proving an intermediate goal in a proof (declared by tactics such as \texttt{have}), the explanation sentences for the proof steps of that proposition become the children of the proof step that declares the intermediate goal.

\begin{figure}[t]
	\centering
	\includegraphics[width=7.5cm]{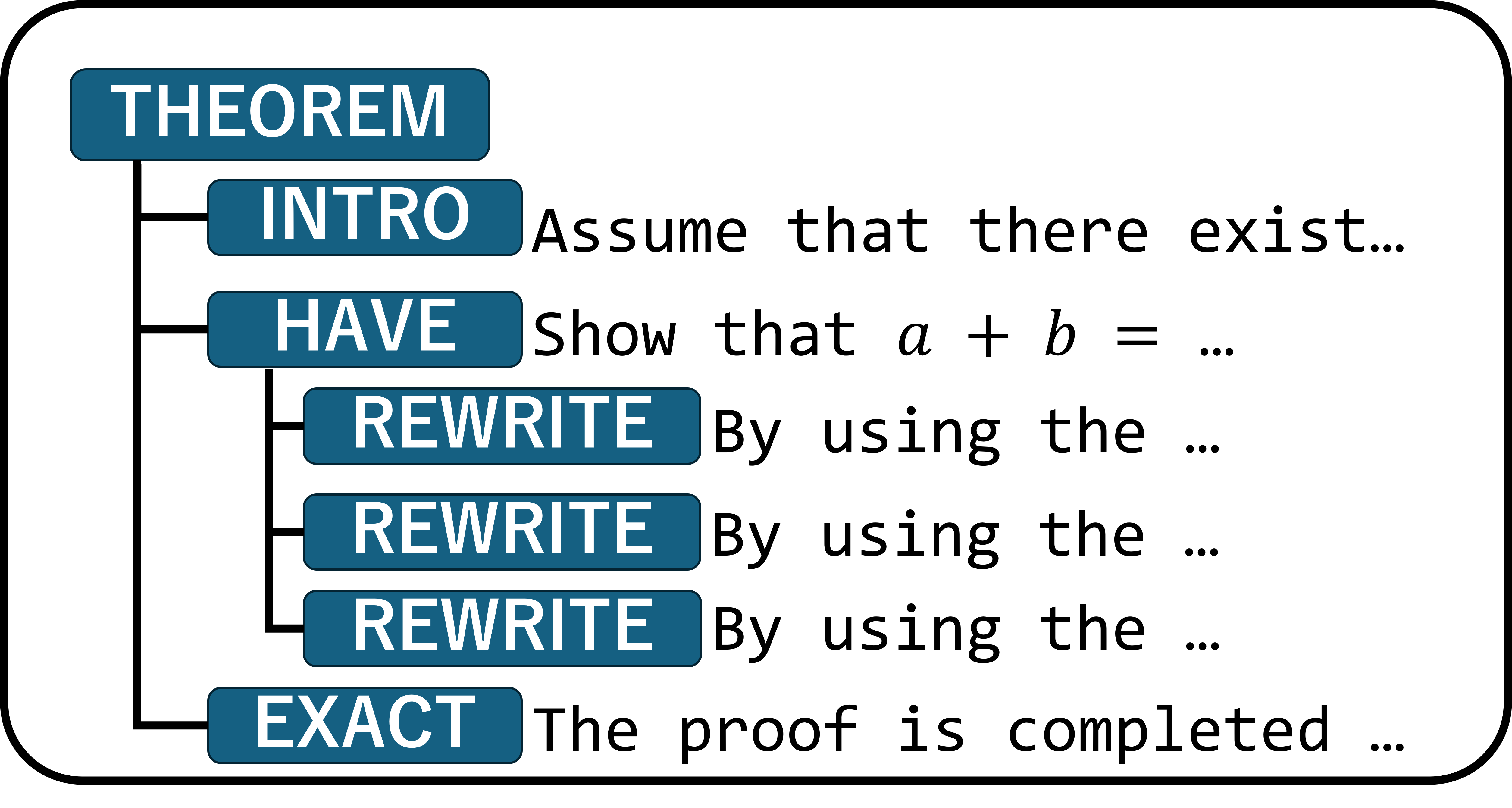}
	\caption{Dependencies among proof steps}\label{fig:parse-dependency}
\end{figure}

Figure \ref{fig:parse-dependency} shows an example of the tree structure. It represents the structure of the formal proof shown in Figure \ref{ex:FL-Proof}. Here, four types of tactics are used: \texttt{intro}, \texttt{have}, \texttt{rw}, and \texttt{exact}. Of these, the three \texttt{rw} tactics are used to prove the intermediate goal declared by the \texttt{have} tactic. Therefore, the dependency structure is represented by a subtree where the explanation sentence for the \texttt{have} tactic is the root, and the explanation sentences for the three \texttt{rw} tactics are the leaves.

\subsection{Summarization of Informalized Steps}\label{method-summarization}

In this step, we leverage the dependency structure tree constructed in the previous step to summarize the operation explanations of all proof steps into a single proof.

First, we use an LLM to informalize the proposition to be proven (the proposition declared at the beginning of the formal proof), which is the root of a tree structure. For this informalization generation, we use a different prompt than the one employed in \ref{method-informalization}, including three manually created few-shot examples, and explanations of definitions used in the proposition.

Next, to generate a natural language proof that aligns with the structure of the formal proof, we generate a sub-proof for each subtree in the tree structure. A sub-proof is the natural language proof of an intermediate goal declared by the tactic at the root of a subtree, and its content is created from the operation explanation sentences corresponding to the nodes of the subtree. Furthermore, the sub-proofs are recursively generated. That is, if a subtree contains further subtrees, the sub-proofs of the contained subtrees are generated first, and they are summarized to the sub-proof of the original subtree. This process is repeated until we get a summary of the whole proof.
\section{Experiment}
\subsection{Experiment Settings}\label{experiment:setting}

For the step-wise informalization and summarization, we used OpenAI's GPT-4.1-mini (gpt-4.1-mini-2025-04-14). In the informalization task, the generation temperature was set to 0.4 to obtain accurate output that adheres to the format of the template and few-shot examples. For the summarization task, to generate natural proof texts, the generation temperature was set to 1.0. 

For evaluating the proposed method, we used 17 proofs from Sections 1.1 and 1.2 of "Calculus I - Calculus of one variable -" by Shizuo Miyajima \shortcite{miyajima}, which were manually formalized according to the structure of the original proofs.

\begin{table*}[t]
\centering
\begin{tabular}{|cc||c|c|c|c|c|c|}
\hline
w/ template & \parbox[t]{0.22\columnwidth}{w/ premise\\ \null\quad\; library} & Correct & Misinformation & \parbox[t]{0.25\columnwidth}{Insufficient\\Information} & \parbox[t]{0.25\columnwidth}{Unnecessary\\Mention} & \parbox[t]{0.25\columnwidth}{Untranslated\\Expression} \\ \hline
$\checkmark$ & $\checkmark$ & \bf{89.05} & 5.15 & \phantom{1}\bf{1.50} & \phantom{1}3.87 & 0.40\\ \hline
$\checkmark$ &  & 83.09 & 8.05 & \phantom{1}4.51 & \phantom{1}\bf{3.46} & 0.89\\ \hline
 & $\checkmark$ & 53.95 & \bf{2.50} & 18.80 & 24.40 & 0.40\\ \hline
 &  & 53.22 & 3.54 & 18.44 & 24.48 & \bf{0.32}\\ \hline
\end{tabular}
\caption{Informalization Evaluation Results. The numbers in the table represent percentages relative to the total number of evaluated proof steps.}
\label{table:informalize_eval}
\end{table*}
\begin{figure}[t!]
\centering
\begin{screen}
  \parindent0pt
  \small
\begin{algorithmic}

\Require Informalization result of proof step $s$
\If {$s$ contains output written in formal language syntax}
 \State \textbf{return} "Untranslated Expression"
\ElsIf {$s$ contains logical/content errors}
 \State \textbf{return} {"Misinformation"}
\ElsIf {$s$ is unclear/lacks sufficient information}
 \State \textbf{return} "Insufficient Information"
\ElsIf {$s$ contains redundant/inappropriate expressions\\ \qquad\; for an explanation}
 \State \textbf{return} "Unnecessary Mention"
\Else
 \State \textbf{return} "Correct"
\EndIf
\end{algorithmic}
\end{screen}
\caption{Evaluation Procedure}
\label{alg:informalize-evaluation}
\end{figure}
\subsection{Evaluation of Step-wise Informalization}
The evaluation of the informalization output was conducted on 1,242 informalized proof steps created by applying the proposed method to a total of 38 formal proofs, which include the evaluation data mentioned in \ref{experiment:setting} plus 21 lemmas used in the formal proofs. The lemmas include theorems taken from the same textbook (Miyajima, 2010), as well as helper lemmas used in the formalization of the proofs (e.g., commutativity of conjunctions). Additionally, to confirm the effects of the templates and premise library, four types of generation were prepared by toggling the presence/absence of the templates and the premise library.

The output was evaluated along the following four dimensions.

\begin{itemize}
\item \textbf{Accuracy of Included Information}\\
Evaluates whether the mathematical expressions and operations appearing in the text correctly reflect the content of the proof step.
\item \textbf{Sufficiency of Included Information}\\
Evaluates whether the generated explanation sufficiently contains the necessary information to describe the proof step.
\item \textbf{Necessity of Included Information}\\
Evaluates whether the explanation contains only necessary information.
\item \textbf{Appropriateness of Translation}\\
Evaluates whether the output is an appropriate natural language explanation that does not include expressions in formal proof syntax.
\end{itemize}
The evaluation was performed manually by the first authors, who holds a degree in applied mathematics, following the procedure shown in Figure \ref{alg:informalize-evaluation}. The evaluation order in the procedure was set based on the degree of the negative impact of the defects on the summarization result.

The experimental results are shown in Table~\ref{table:informalize_eval}. 
The results show that the proposed method achieves the most accurate informalization when both the templates and the premise library are used. When templates are present, the proposed method achieves an accuracy over 80\%, but without templates, the accuracy drops to approximately 50\%. Counterintuitively, Misinformation increases when the templates are used; this is because necessary content such as references to specific variables and hypotheses, and reference to an already-proven theorem is significantly shorter when the templates are not used, consequently decreasing incorrect information.
We further validated these findings using McNemar’s test ($\alpha=0.05$). The results show that all pairwise comparisons yielded significant differences except for the comparison between w/o template + w/o premise library and w/o template + w/ premise library, where no significant improvement was observed. Detailed results are provided in Appendix C.

In the following, we scrutinize the effect of premise library and templates.

\subsubsection*{Effects of premise library}
Using the premise library in conjunction with the templates improves accuracy by approximately 6\%. This is primarily due to the supression of Misinformation and Insufficient Information, suggesting that providing premises, i.e., natural language explanation of definitions and theorems mentioned in a proof, makes it easier to properly refer to them. On the other hand, when templates are absent, adding the premise library shows little change in accuracy. This is presumably because the premise library synergistically improves performance in tandem with template-based generation by allwing the LLM to fill the slots simply by directly quoting the natural language description of the premises.

\subsubsection*{Effects of templates}
Comparing the results with and without templates reveals a significant increase in insufficient information and unnecessary mentions when templates are absent. This indicates that while the templates function effectively as constraints on the output, few-shot examples alone are insufficient to control the output. Templates determine the format of the output, making it less likely for superfluous information to be generated while ensuring that necessary information is included.
\subsubsection*{Untranslated formal expressions}
Formal language expressions are occasionally outputted directly in the translation and its frequency does not change significantly with or without templates or the premise library. This is caused by formal language definitions (e.g., expressing the infimum of a set as \texttt{sInf}) and notation specific to formal languages (e.g., expressing the $n$-th element of a sequence $x$ as \texttt{x n}) appearing directly in the output. While the former problem is somewhat ameliorated by the premise library, the latter problem is due to the shortage of the LLM's internal knowledge of mathematical notation and formal language notation. Thus, no significant improvement was observed.


\subsection{Evaluation of Summarization}
The evaluation of summarization result was conducted using the 17 paired data of formal and natural language proofs mentioned in \ref{experiment:setting}. First, for each of the 17 natural language proofs, an average of 6.4 evaluation criteria were set. Each evaluation criterion included key points that should be mentioned in the proof, such as the proposition to be proven, theorems used, and newly introduced variables. These evaluation criteria were all manually set based on the original natural language proofs. 

We applied the proposed method to the formal proof to generate a natural language proof, and evaluated whether the result correctly mentioned each evaluation criterion on a three-point scale. The evaluation criteria were: "key points correctly captured (\coloremoji{😄})", "key points partially captured (\coloremoji{🤨})", and "key points not captured (\coloremoji{☹️})" (detailed evaluation criteria are provided in Appendix D). The score for the generated result was then calculated as follows:
\[
  \text{score} = \frac{\mbox{count}(\coloremoji{😄}) + 0.5 \times \mbox{count}(\coloremoji{🤨})}{\mbox{count}(\coloremoji{😄})+\mbox{count}(\coloremoji{🤨})+\mbox{count}(\coloremoji{☹️})}.
\]
To verify the effectiveness of recursive summarization, we generated two sets of summaries, with and without the dependency structure analysis and recursive summarization.

\begin{table}[t]
\centering
\begin{tabular}{|l||c|c|c|c|}
\hline Human Evaluation & \coloremoji{😄} & \coloremoji{🤨} & \coloremoji{☹️} & Score \\ \hline
w/ recursive & 87 & 11 & 10 & 0.857\\ \hline
w/o recursive & 85 & 10 & 13 & 0.833\\ \hline
\end{tabular}
\caption{Summarization Evaluation Results.}
\label{table:summerize-eval}
\end{table}
The evaluation results are shown in Table \ref{table:summerize-eval}.
The numbers in the table represent the total evaluation counts for the 17 proofs. The score in the table is calculated using the previously mentioned formula, with the denominator being the total number of evaluation criteria across all the 17 proofs.

The results show that recursive summarization captures slightly more key points than the non-recursive approach. Close inspection revealed that the proofs generated using recursive summarization did not exhibit any significant deviations from the original formal proofs or logical breakdowns. In contrast, among the 17 outputs generated without recursive summarization, four included reasoning not in the original proof and/or contained substantial logical inconsistencies. These issues stem from the LLM's inability to fully capture the structure of the original formal proof.

The effect of recursive summarization can be explained as follows. First, the limited input window of the LLM likely contributes to hallucinations. In the recursive summarization approach, summaries are generated for semantically coherent units, keeping input size small and thereby reducing the likelihood of hallucinations. In contrast, without recursive summarization, all proof steps are fed into the model at once, resulting in much larger inputs and increased the risk of information loss and hallucination. These observations suggest that recursive summarization is effective for generating accurate summaries. 
\begin{figure}[t]
\begin{screen}
  \fontsize{10pt}{11pt}\selectfont
\parindent0pt
\fontsize{10pt}{11pt}\selectfont
We will show that the greatest lower bound (infimum) of the set $ P $ of all positive real numbers is zero. 
First, note that $ P $ is nonempty since it contains the number 1, which clearly satisfies $1 > 0$. 
Next, we observe that 0 is a lower bound for $ P $ because every element $ a \in P $ satisfies $ a > 0 $, and thus $ a \geq 0 $.
Therefore, $ P $ is bounded below by 0. Since the infimum is the greatest of all lower bounds, it follows that the infimum of $ P $ is at least zero. To show that the infimum is not greater than zero, suppose by contradiction that $\inf P > 0$.
Then the number $\frac{\inf P}{2}$ is positive and thus belongs to $ P $, but since $\inf P$ is a lower bound, it must satisfy $\frac{\inf P}{2} \geq \inf P$, implying $\inf P \leq 0$, a contradiction.
Hence, $\inf P \leq 0$. Combining these inequalities, we have $\inf P \geq 0$ and $\inf P \leq 0$, which together imply $\inf P = 0$. This completes the proof that zero is the greatest lower bound of the set of all positive real numbers.
\end{screen}
\caption{The proof of ``$\inf \{x \in \mathbb{R} | x > 0\} = 0$`` generated with using recursive summarization.}
\label{ex:with-recursive-summarization}
\end{figure}
\begin{figure}[tbp]
\begin{screen}
  \fontsize{10pt}{11pt}\selectfont
\parindent0pt
We will show that the set $ P $ of all positive real numbers is nonempty, bounded below by zero, and that its greatest lower bound (infimum) is exactly zero.
First, we observe that $ P $ is nonempty since the number 1 belongs to $ P $ by virtue of satisfying the defining property $ 1 > 0 $.
Next, we demonstrate that zero is a lower bound of $ P $ by noting that for any element $ a \in P $, we have $ a > 0 $ and thus $ a \geq 0 $.
Since $ P $ is nonempty and bounded below, the greatest lower bound $ \inf P $ exists.
We then argue by contradiction to show that $ \inf P \leq 0 $: assuming $ \inf P > 0 $, it follows that half of this infimum, $ \frac{\inf P}{2} $, is an element of $ P $ because it is strictly greater than zero.
\hl{However, this contradicts the property of the infimum being a greatest lower bound since $ \frac{\inf P}{2} < \inf P $.}
Therefore, the assumption $ \inf P > 0 $ fails, and we conclude that $ \inf P \leq 0 $. Coupled with the prior conclusion that $ \inf P \geq 0 $, it follows that $ \inf P = 0 $, completing the proof.
\end{screen}
\caption{The proof of ``$\inf \{x \in \mathbb{R} | x > 0\} = 0$'' generated without using recursive summarization.}
\label{ex:without-recursive-summarization}
\end{figure}
As an example, we present a generated proof for the proposition ``the infimum of the set of positive real numbers is $0$''.
The original proof is as follows:
\begin{screen}
  \fontsize{10pt}{11pt}\selectfont
\parindent0pt
In this proof, let $P := \{x \in \mathbb{R} | x > 0\}$.
By definition, 0 is a lower bound of $P$, so $\inf P \ge 0$.
If $\inf P > 0$, then $1/2 \inf P \in P$, so $1/2 \inf P \ge \inf P$ must hold.
This implies $\inf P \le 0$, which contradicts our assumption that $\inf P > 0$.
\end{screen}

Figure \ref{ex:with-recursive-summarization} and \ref{ex:without-recursive-summarization} present the proofs generated with and without using the formal proof structure, respectively.

Comparing these examples, the proof in Figure \ref{ex:with-recursive-summarization} captures the flow of the original proof, though it includes some redundant content. On the other hand, the proof in Figure \ref{ex:without-recursive-summarization} fails to mention that $\inf P \ge 0$, and furthermore, deviates from the logical flow of the original proof in the reasoning expressed in the highlighted sentence.
\subsection{Translation of Mathlib Formal Proofs}
\begin{figure}[tbp]
\begin{lstlisting}[mathescape, numbers=none]
  theorem exists_infinite_primes (n : $\mathbb{N}$) :
    $\exists$ p, n $\leq$ p $\wedge$ Nat.Prime p := by
    let p := minFac (n ! + 1)
    have f1 : n ! + 1 $\neq$ 1 := 
      Nat.ne_of_gt  <| succ_lt_succ  
                    <| factorial_pos _
    have pp : Nat.Prime p := 
      minFac_prime f1
    have np : n $\leq$ p :=
      le_of_not_ge fun h =>
        have h$_1$ : p $\mid$ n ! := 
          dvd_factorial (minFac_pos _) h
        have h$_2$ : p $\mid$ 1 := 
          (Nat.dvd_add_iff_right h$_1$).2 
            (minFac_dvd _) 
            pp.not_dvd_one h$_2$
    exact $\langle$p, np, pp$\rangle$
  \end{lstlisting}
  \caption{The formal proof of ``there are infinitely many prime numbers.'' in Mathlib.}
  \label{proof:formal_prime_numbers}
\end{figure}
\begin{figure}[tbp]
  \begin{screen}
    \fontsize{10pt}{11pt}\selectfont
  We will show that for any natural number \( n \), there exists a prime number \( p \) such that \( p \geq n \). Consider the number \( n! + 1 \). Since the factorial \( n! \) is always positive for natural \( n \), it follows that \( n! + 1 > 1 \), so \( n! + 1 \neq 1 \). Let \( p \) be the smallest prime factor of \( n! + 1 \). By definition, \( p \) is prime, as the smallest prime factor of any integer greater than 1 must itself be prime. Next, suppose for the sake of contradiction that \( p < n \). Since \( p \leq n \), \( p \) divides \( n! \) by the definition of factorial. Furthermore, \( p \) divides \( n! + 1 \) by choice. Therefore, \( p \) divides the difference \((n! + 1) - n! = 1\), which is impossible since no prime number divides 1. This contradiction implies that \( p \geq n \). Hence, for any natural number \( n \), there exists a prime \( p \) with \( p \geq n \), completing the proof.
  \end{screen}
  \caption{The generated proof of ``there are infinitely many prime numbers.''}
  \label{fig:proof_prime_numbers}
\end{figure}
The data evaluated so far consisted of existing natural language proofs, for which formal proofs were then developed. In contrast, this section presents examples of natural language proofs generated from Mathlib's formal proofs using our proposed method. 

As an example, we will use the proposition ``there are infinitely many primes''. The original formal proof of the theorem is presented in Figure \ref{proof:formal_prime_numbers}. This proceeds as follows:
\begin{itemize}
  \item Let $p$ be the smallest prime factor of $(n! + 1)$.
  \item That $n! + 1 \mid 1$ since factorial is always positive.
  \item If $n > p$, then from the property of factorials, $p \mid n!$ and $p \mid n! + 1$ holds, then it also holds that $p \mid 1$. However, since prime numbers do not divide 1, this is a contradiction. Therefore, $n \leq p$ holds.
\end{itemize}
The informalize result is shown in Figure \ref{fig:proof_prime_numbers}. 
This demonstrates that even when formal proofs are not created to directly follow human-written proofs, it is possible to generate proofs that are still quite readable as human-written ones. Several other examples are given in Appendix E.

\section{Conclusion}

We proposed a natural language translation method for formal proofs by informalizing each proof step with template and premise library, and summarizing the result recursively, thereby generating human-readable natural language proofs. We applied the proposed method to formal proofs formalized according to university-level mathematical proofs and demonstrated that it can generate readable proofs that capture 86\% of the key points. We also showed that the proposed method can generate high-quality natural language proofs from Mathlib's formal proofs, i.e., even from those without corresponding natural language proofs.
\paragraph*{Supplementary Materials Availability Statement:}
The appendix will be submitted as a PDF file via the submission form. The source code and output examples is publicly available at \url{https://github.com/hattori-matsuzakilab/AutoInformalizationWithTemplate}.
\paragraph*{Acknowledgments:}This research was supported in part by KIOXIA Corporation.
\clearpage
\bibliography{references}
\clearpage
\appendix
\section{Limitation}
\subsection{Construction of Templates and Few-shot Examples}
The step-by-step informalization of formal proofs proposed in this research only becomes effective when a unique template is prepared for every possible operation executable by a tactic, along with multiple few-shot examples for each tactic. However, identifying all possible operations a tactic can perform and creating appropriate few-shot examples for them is extremely time-consuming.

To make the method proposed in this research applicable to a wider range of formal proofs, it is desirable to implement tools that assist in the creation of templates and few-shot examples, or an automatic generation tool utilizing LLMs.
\subsection{Regarding the Summarization Generation Method}
The summarization method proposed in this research aims to generate natural text by recursively summarizing based on dependency structures. This is intended to act like a "sieve," repeatedly selecting information necessary for the LLM to construct natural proof texts, thereby filtering out trivial information. Therefore, the more complex the proof structure is, the more redundant explanations are removed, bringing the generated proof closer to a human-written proof. However, a challenge arises when dealing with simple proofs that contain trivial operations not explicitly stated in human-written proofs: the number of times the text is sieved decreases, making it more likely for details that are trivial for human to be outputted.

To solve this problem, we believe it is necessary to construct a method that explicitly marks important information and reference expressions to be retained during summarization, rather than relying on a simple recursive algorithm.
\section{Informalization Prompt} \label{appendix:informalize_prompt}
    \begin{breakbox}
    \parindent0pt
    \small
You are an expert in Lean4 and formal mathematics. \\
Transform the given Lean4 tactic into a clear and concise natural language explanation, accurately conveying the operation performed as a step in a mathematical proof without using the format of the formal language.\\
Ensure that any predicates from the formal language are not included in the explanation.\\
\# Steps\\
1. **Understand the Applied Tactic**:\\
- Analyze the `Applied Tactic` to comprehend its function within the proof.\\
\null\quad	- If a tactic involves using one or more theorems or definitions, read the theorems listed under `Using Definitions and Theorems` and ensure you understand their content.\\
2. **Examine Hypotheses and Goals**:\\
- Compare `Hypotheses And Goals Before Tactic Application` with `Hypotheses And Goals After Tactic Application` to understand what changes in objectives occurred.\\
\null\quad    - Hypotheses and goals are separated with the symbol '$\vdash$'.\\
3. **Formulate the Explanation**:\\
- Describe what action the tactic took, referring only to what was altered based on the changes observed in the hypotheses and goals.\\
- Avoid directly referencing predicates from the formal language.\\
\null\quad    - When referring to variables, be sure to explicitly use these variable names in the output.\\
\null\qquad        - (ex) Set $ A $ is Empty.\\
\null\quad    - **Do not** include the names of theorems or definitions in formal languages, or variables used as aliases to specific expressions (like h : x = 2 * y) in the output. Instead, explain the content in natural language.\\
\# Output Format\\
- Provide a natural language explanation summarizing the operation of the tactic and the immediate effects on the hypotheses and goals.\\
\null\quad  - Provide all necessary information for the explanation in precise and detailed terms.\\
\# Examples\\
\#\#\# Example X\\
**Input Information**:\\
- Applied Tactic: [example.appliedTactic]\\
- Hypotheses And Goals Before Tactic Application: [example.goalsBefore]\\
- Hypotheses And Goals After Tactic Application: [example.goalsAfter]\\
**Using Definitions and Theorems**:\\
- [example.premises]
**Output**:\\
- [example.output]\\
\# Notes\\
- Always determine what assumptions or definitions are brought into effect or altered.\\
- Make sure explanations are precise to maintain clarity and avoid unnecessary details.\\
- Do not include expressions written in Lean's formal language in the output.  \\
\null\quad  - In formal proofs, casts such as from natural numbers to integers are represented by $ \uparrow $. Therefore, ensure that the output does not contain $ \uparrow $ representing a cast.\\
- If you use explanations or citations to output formulas, please use TeX formatting for citations if the formulas are not complex, or use explanations written in natural language if the formulas are complex.\\
Using the example above as a reference, please explain the following input in natural language as one operation in a mathematical proof.\\
**Input Information**:\\
- Applied Tactic: [input.appliedTactic]\\
- Hypotheses And Goals Before Tactic Application: [input.goalsBefore]\\
- Hypotheses And Goals After Tactic Application: [input.goalsAfter]\\
**Using Definitions and Theorems**:\\
- [input.premises]
**Output**:
\end{breakbox} 
\section{Results of McNemar's test for the tabulated results of the inverse formalization}
\begin{table*}[t]
\centering
\small
\begin{tabular}{|c|c||c|c|c|c|}
\hline
method A & method B & A only count & B only count & $\chi^2$ & p-value \\ \hline
\parbox[t]{0.22\columnwidth}{w/ template\\ w/o library} & \parbox[t]{0.22\columnwidth}{w/ template\\ w/ library} & 51 & 123 & \phantom{1}28.9713 & 7.3460e-08 \\ \hline
\parbox[t]{0.22\columnwidth}{w/o template\\ w library} & \parbox[t]{0.22\columnwidth}{w/ template\\ w/ library} & 40 & 475 & 365.7398 & 1.5841e-81 \\ \hline
\parbox[t]{0.22\columnwidth}{w/o template\\ w library} & \parbox[t]{0.22\columnwidth}{w/ template\\ w/o library} & 56 & 418 & 274.9388 & 9.5181e-62 \\ \hline
\parbox[t]{0.22\columnwidth}{w/o template\\ w/o library} & \parbox[t]{0.22\columnwidth}{w/ template\\ w/ library} & 42 & 477 & 362.9210 & 6.5096e-81 \\ \hline
\parbox[t]{0.22\columnwidth}{w/o template\\ w/o library} & \parbox[t]{0.22\columnwidth}{w/ template\\ w/o library} & 62 & 423 & 267.2165 & 4.5875e-60 \\ \hline
\parbox[t]{0.22\columnwidth}{w/o template\\ w/o library} & \parbox[t]{0.22\columnwidth}{w/o template\\ w/ library} & 115 & 118 & \phantom{2}0.0172 & 8.9576e-01 \\ \hline
\end{tabular}
\caption{Results of McNemar's test for the tabulated results of the informalization. In the table, "library" refers to the Premise Library. "A only count" and "B only count" indicate the number of cases where the result was correct by method A but incorrect by method B, and incorrect by method A but correct by method B, respectively. $\chi^2$ denotes the test statistic.}
\label{table:mcnemar-informalize}
\end{table*}
Table \ref{table:mcnemar-informalize} shows the results of McNemar’s test on the informalization results, conducted at $\alpha = 0.05$. First, when comparing methods without / with templates, the number of correct results is clearly higher when templates are applied, and the corresponding p-value is small, indicating a statistically significant difference. Regarding the Premise Library, a significant difference is observed only when templates are applied, while no significant difference is found when templates are not used. This indicates that the Premise Library is effective only in combination with templates.
\section{Criteria for summarization evaluation}
For the proof summarization evaluation, "key points" were identified by exhaustively extracting items from the original natural language proof that fall into the following categories:
\begin{enumerate}
    \item Definitions of variables and constants used in the proof.
    \item Statements of propositions corresponding to sub-lemmas.
    \item Proofs of the sub-lemmas.
    \item Mentions of proof methods (e.g., proof by contradiction).
    \item References to theorems or lemmas used.
\end{enumerate}
Categories 2 and 3 may be combined with 4 and 5, for instance in cases such as ``mentioning the use of proof by contradiction to prove A.'' 

Each item was then evaluated according to the following criteria:
\begin{itemize}
    \item \textbf{Correctly captured}: The item was mentioned correctly.
    \item \textbf{Partially captured}: The mention was correct but incomplete.
    \item \textbf{Not captured}: The item was omitted or the reasoning was incorrect.
\end{itemize}

For composite categories, an item was rated as correctly captured only if all component criteria were satisfied.

\section{Generation Example}\label{appendix:informalize}
\subsection{Example 1}
Theorem : The interior of a closed ball centered at a point \( x \) with radius \( r \) in a metric space is precisely the open ball centered at \( x \) with the same radius \( r \).
\begin{lstlisting}[mathescape,numbers=none]
theorem interior_closedBall (x : E) {r : $\mathbb{R}$} (hr : r $\neq$ 0) :
  interior (closedBall x r) = ball x r := by
  cases' hr.lt_or_lt with hr hr
  . rw [closedBall_eq_empty.2 hr, ball_eq_empty.2 hr.le, interior_empty]
  refine Subset.antisymm ?_ ball_subset_interior_closedBall
  intro y hy
  rcases (mem_closedBall.1 <| interior_subset hy).lt_or_eq with (hr|rfl)
  . exact hr
  set f : $\mathbb{R}$ $\rightarrow$ E := fun c : $\mathbb{R}$ => c $\cdot$ (y - x) + x
  suffices f $^{-1'}$ closedBall x (dist y x) $\subseteq$ Icc (-1) 1 by
    have hfc : Continuous f := 
      (continuous_id.smul continuous_const).add continuous_const
    have hf1 : (1 : $\mathbb{R}$) $\in$ f $^{-1'}$ interior (closedBall x <| dist y x) := by simpa [f]
    have h1 : (1 : $\mathbb{R}$) $\in$ interior (Icc (-1 : $\mathbb{R}$) 1) := interior_mono
      this (preimage_interior_subset_interior_preimage hfc hf1)
    simp at h1
  intro c hc
  rw [mem_Icc, $\leftarrow$ abs_le, $\leftarrow$ Real.norm_eq_abs, $\leftarrow$ mul_le_mul_right hr]
  simpa [f, dist_eq_norm, norm_smul] using hc
\end{lstlisting}
\begin{breakbox}
\small
We will show that the interior of a closed ball centered at a point \( x \) with radius \( r \) in a metric space is precisely the open ball centered at \( x \) with the same radius \( r \). \\
First, consider the case when \( r < 0 \). By the definition of a closed ball, no points satisfy the distance condition for a negative radius, so both the closed ball and the open ball are empty sets. Since the interior of the empty set is empty, the proposition holds trivially in this case.\\
Next, for \( r \geq 0 \), we prove the equality by showing mutual inclusion. Specifically, it suffices to prove that the interior of the closed ball is contained within the open ball. Assume \( y \) is an element of the interior of the closed ball centered at \( x \) with radius \( r \). To establish that \( y \) lies in the open ball, we conduct a case analysis on the distance \( d(y,x) \). If \( d(y,x) < r \), then \( y \) is clearly in the open ball by definition, so the claim holds.\\
To treat the remaining case, we define a function \( f:\mathbb{R} \to E \) by \( f(c) = c \cdot (y - x) + x \). This function is continuous, as it is composed of scalar multiplication and vector addition, both of which preserve continuity. Since \( y \) lies in the interior of the closed ball with radius \( d(y,x) \), it follows that the scalar 1 (corresponding to \( y = f(1) \)) is an interior point of the preimage under \( f \) of this interior. Given that the preimage of the closed ball under \( f \) is contained in the interval \([-1, 1]\), this implies that 1 is an interior point in \([-1, 1]\). Using the continuity of \( f \) and properties of norms and distances, we show that any \( c \) such that \( f(c) \) lies in the closed ball must satisfy \( c \in [-1,1] \). By exploiting the positivity of \( d(y,x) \) and the equivalence of norms and absolute values, we verify the inequalities establishing this containment. Together, these arguments imply that \( y \) must lie within the open ball centered at \( x \) with radius \( r \), completing the proof that the interior of the closed ball equals the open ball with the same parameters.
\end{breakbox}
\subsection{Example 2}
Theorem : Heron's formula
\begin{lstlisting}[mathescape,numbers=none]
theorem heron {p1 p2 p3 : P} (h1 : p1 $\neq$ p2) (h2 : p3 $\neq$ p2) :
    let a := dist p1 p2
    let b := dist p3 p2
    let c := dist p1 p3
    let s := (a + b + c) / 2
    1 / 2 * a * b * sin ($\angle$ p1 p2 p3) = $\sqrt{}$ (s * (s - a) * (s - b) * (s - c)) := by
  intro a b c s
  let $\gamma$ := $\angle$ p1 p2 p3
  obtain := (dist_pos.mpr h1).ne', (dist_pos.mpr h2).ne'
  have cos_rule : cos $\gamma$ = (a * a + b * b - c * c) / (2 * a * b) := by
    field_simp [mul_comm,
    dist_sq_eq_dist_sq_add_dist_sq_sub_two_mul_dist_mul_dist_mul_cos_angle p1 p2 p3]
  let numerator := (2 * a * b) ^ 2 - (a * a + b * b - c * c) ^ 2
  let denominator := (2 * a * b) ^ 2
  have split_to_frac : $\uparrow$1 - cos $\gamma$ ^ 2 = numerator / denominator := by field_simp [cos_rule]
  have numerator_nonneg : 0 $\leq$ numerator := by
    have frac_nonneg : 0 $\leq$ numerator / denominator :=
      (sub_nonneg.mpr (cos_sq_le_one $\gamma$)).trans_eq split_to_frac
    cases' div_nonneg_iff.mp frac_nonneg with h h
    . exact h.left
    . simpa [numerator, denominator, a, b, c, h1, h2] using le_antisymm h.right (sq_nonneg _)
  have ab2_nonneg : 0 $\leq$ 2 * a * b := by positivity
  calc
    1 / 2 * a * b * sin $\gamma$ = 1 / 2 * a * b * ($\sqrt{}$ numerator / $\sqrt{}$ denominator) := by
      rw [sin_eq_sqrt_one_sub_cos_sq, split_to_frac, sqrt_div numerator_nonneg] <;>
        simp [$\gamma$, angle_nonneg, angle_le_pi]
    _ = 1 / 4 * $\sqrt{}$ ((2 * a * b) ^ 2 - (a * a + b * b - c * c) ^ 2) := by
      field_simp [numerator, denominator, ab2_nonneg]; ring
    _ = $\uparrow$1 / $\uparrow$4 * $\sqrt{}$ (s * (s - a) * (s - b) * (s - c) * $\uparrow$4 ^ 2) := by simp only [s]; ring_nf
    _ = $\sqrt{}$ (s * (s - a) * (s - b) * (s - c)) := by
      rw [sqrt_mul', sqrt_sq, div_mul_eq_mul_div, one_mul, mul_div_cancel_right$$_0] <;> norm_num
\end{lstlisting}
\begin{breakbox}
\small
We will prove Heron's formula by showing that half the product of two sides of a triangle and the sine of their included angle equals the square root of the semiperimeter times its differences with each side. Let \(a\), \(b\), and \(c\) be the lengths of the sides opposite points \(p_1\), \(p_3\), and \(p_2\), respectively, and let \(s = \frac{a + b + c}{2}\) denote the semiperimeter. Define \(\gamma\) as the angle formed at vertex \(p_2\) between sides \(a\) and \(b\). Noting that \(a\) and \(b\) are positive since the points are distinct, we proceed to express \(\cos \gamma\) via the Law of Cosines as \(\cos \gamma = \frac{a^2 + b^2 - c^2}{2ab}\). Squaring this expression and subtracting from 1, we write \(1 - \cos^2 \gamma\) as a ratio of polynomial expressions in \(a, b, c\), specifically \(\frac{\text{numerator}}{\text{denominator}}\), where the numerator is defined as the difference between \((2ab)^2\) and \((a^2 + b^2 - c^2)^2\), and the denominator is \((2ab)^2\). We establish that this numerator is nonnegative by analyzing the quotient and considering the range of \(\cos^2 \gamma\), then employing a case analysis of the signs of numerator and denominator along with the nonnegativity of squares and inequalities among the side lengths. Furthermore, we show that \(2ab\) is nonnegative since it appears multiplied by \(\sin \gamma\), which corresponds to geometric quantities bounded between zero and \(\pi\). Confirming that the angle \(\gamma\) satisfies \(0 \leq \gamma \leq \pi\), we express \(\sin \gamma\) as \(\sqrt{1 - \cos^2 \gamma}\) and simplify the corresponding expressions by substituting the polynomial ratios and simplifying via algebraic equivalences. Through these steps, we verify the key equality \(\frac{1}{2} ab \sin \gamma = \sqrt{s(s-a)(s-b)(s-c)}\), completing the proof of Heron's formula as desired.
\end{breakbox}
\subsection{Example 3}
Theorem: $f$ and $g$ are real-valued functions defined on the real line. For all $x$ and $y$,
$f(x + y) + f(x - y) = 2f(x)g(y)$. $f$ is not identically zero and $|f(x)| \leq 1$ for all $x$.
Prove that $|g(x)| \leq 1$ for all $x$. (IMO 1972 Q5)
\begin{lstlisting}[mathescape,numbers=none]
theorem imo1972_q5 (f g : $\mathbb{R} \rightarrow \mathbb{R}$) (hf1 : $\forall$ x, $\forall$ y, f (x + y) + f (x - y) = 2 * f x * g y)
  (hf2 : $\forall$ y, $\|$f y$\|$ $\leq$ 1) (hf3 : $\exists$ x, f x $\neq$ 0) (y : $\mathbb{R}$) : $\|$g y$\|$ $\leq$ 1 := by
-- Suppose the conclusion does not hold.
by_contra! hneg
set S := Set.range fun x => $\|$f x$\|$
-- Introduce `k`, the supremum of `f`.
let k : $\mathbb{R}$ := sSup S
-- Show that `$\|$f x$\|$ $\leq$ k`.
have hk$_1$ : $\forall$ x, $\|$f x$\|$ $\leq$ k := by
  have h : BddAbove S := $\langle$1, Set.forall_mem_range.mpr hf2$\rangle$
  intro x
  exact le_csSup h (Set.mem_range_self x)
-- Show that `2 * ($\|$f x$\|$ * $\|$g y$\|$) $\leq$ 2 * k`.
have hk$_2$ : $\forall$ x, 2 * ($\|$f x$\|$ * $\|$g y$\|$) $\leq$ 2 * k := fun x $\mapsto$
  calc
    2 * ($\|$f x$\|$ * $\|$g y$\|$) = $\|$2 * f x * g y$\|$ := by simp [abs_mul, mul_assoc]
    _ = $\|$f (x + y) + f (x - y)$\|$ := by rw [hf1]
    _ $\leq$ $\|$f (x + y)$\|$ + $\|$f (x - y)$\|$ := norm_add_le _ _
    _ $\leq$ k + k := add_le_add (hk$_1$ _) (hk$_1$ _)
    _ = 2 * k := (two_mul _).symm
set k' := k / $\|$g y$\|$
-- Demonstrate that `k' < k` using `hneg`.
have H$_1$ : k' < k := by
  have h$_1$ : 0 < k := by
    obtain $\langle$x, hx$\rangle$ := hf3
    calc
      0 < $\|$f x$\|$ := norm_pos_iff.mpr hx
      _ $\leq$ k := hk$_1$ x
  rw [div_lt_iff]
  . apply lt_mul_of_one_lt_right h$_1$ hneg
  . exact zero_lt_one.trans hneg
-- Demonstrate that `k $\leq$ k'` using `hk$_2$`.
have H$_2$ : k $\leq$ k' := by
  have h$_1$ : $\exists$ x : $\mathbb{R}$, x $\in$ S := by use $\|$f 0$\|$; exact Set.mem_range_self 0
  have h$_2$ : $\forall$ x, $\|$f x$\|$ $\leq$ k' := by
    intro x
    rw [le_div_iff]
    . apply (mul_le_mul_left zero_lt_two).mp (hk$_2$ x)
    . exact zero_lt_one.trans hneg
  apply csSup_le h$_1$
  rintro y' $\langle$yy, rfl$\rangle$
  exact h$_2$ yy
-- Conclude by obtaining a contradiction, `k' < k'`.
apply lt_irrefl k'
calc
  k' < k := H$_1$
  _ $\leq$ k' := H$_2$
\end{lstlisting}
\begin{breakbox}
  \small
  We will show that if a function \( f \) satisfies the functional equation \( f(x + y) + f(x - y) = 2 f(x) g(y) \) for all real numbers \( x, y \), and if \( |f(x)| \leq 1 \) for all \( x \) with some \( |g(y)| > 1 \), then this leads to a contradiction, thereby implying that \( |g(y)| \leq 1 \).

To proceed, let \( S = \{ |f(x)| : x \in \mathbb{R} \} \) denote the set of norms of values of \( f \), and let \( k = \sup S \) be its supremum. Since \( |f(x)| \leq 1 \) for all \( x \), the set \( S \) is bounded above by 1, ensuring \( k \leq 1 \). For any real number \( x \), we have \( |f(x)| \leq k \) by definition of the supremum.

Next, we observe that for all \( x, y \in \mathbb{R} \), the functional equation implies
\begin{align*}
2 |f(x)| |g(y)| &= |2 f(x) g(y)| = |f(x + y) + f(x - y)| \\
                &\leq |f(x + y)| + |f(x - y)| \leq 2k.
\end{align*}
Dividing both sides by \( 2 |g(y)| \), we define
\[
k' := \frac{k}{|g(y)|}.
\]
Now, since \( k \geq |f(x)| > 0 \) for some \( x \) where \( f(x) \neq 0 \), we have \( k > 0 \). Because we assumed \( |g(y)| > 1 \), it follows that
\[
k' = \frac{k}{|g(y)|} < k,
\]
as dividing by a number greater than one reduces the value.

On the other hand, considering the inequality for arbitrary \( x \),
\[
|f(x)| \leq k,
\]
and from the previous inequality,
\[
2 |f(x)| |g(y)| \leq 2k,
\]
which implies
\[
|f(x)| \leq \frac{k}{|g(y)|} = k'.
\]
Since this is true for all \( x \), it follows that every element of \( S \) is bounded above by \( k' \); hence \( k \leq k' \) by definition of supremum.

Combining these two inequalities yields the contradiction
\[
k' < k \leq k',
\]
which cannot hold. This contradiction arises from the assumption that \( |g(y)| > 1 \).

Therefore, we conclude that \( |g(y)| \leq 1 \) for all \( y \), completing the proof.
\end{breakbox}

\end{document}